%% file: main.tex
\documentclass[letterpaper, 10 pt, conference]{ieeeconf}  
\pdfoutput=1

\IEEEoverridecommandlockouts                              

\usepackage{graphics} 
\usepackage{amsmath} 
\usepackage{amssymb}  
\usepackage{graphicx}
\usepackage{hyperref}
\usepackage{subcaption}
\usepackage{graphicx}
\usepackage{mathtools}
\usepackage{dsfont}
\usepackage{float}
\usepackage{makecell}
\usepackage{authblk}
\usepackage{algorithm}
\usepackage{algcompatible}

\renewcommand{\baselinestretch}{1.0}

\makeatletter
\renewcommand\AB@affilsepx{\hspace{1in} \protect\Affilfont}
\makeatother

\newcommand\blfootnote[1]{%
  \begingroup
  \renewcommand\thefootnote{}\footnote{#1}%
  \addtocounter{footnote}{-1}%
  \endgroup
}

\DeclareMathOperator{\E}{\mathbb{E}}

\title{\LARGE \bf
Residual Reinforcement Learning for Robot Control
}

\author{Tobias Johannink$^{*1,3}$, Shikhar Bahl$^{*2}$, Ashvin Nair$^{*2}$, Jianlan Luo$^{1,2}$, Avinash Kumar$^{1}$,\\ Matthias Loskyll$^{1}$, Juan Aparicio Ojea$^{1}$, Eugen Solowjow$^{1}$, Sergey Levine$^2$}

\begin{document}
\maketitle
\blfootnote{$\;^*$ First three authors contributed equally, $\;^1$ Siemens Corporation, \\ $^2$ University of California, Berkeley, $\;^3$ Hamburg University of Technology.}

\thispagestyle{empty}
\pagestyle{empty}

\vspace{-10pt}

\begin{abstract}
Conventional feedback control methods can solve various types of robot control problems 
very efficiently by capturing the structure with explicit models, such as rigid body equations of motion.
However, many control problems in modern manufacturing deal with contacts and friction, which are difficult to capture with first-order physical modeling. 
Hence, applying control design methodologies to these kinds of problems often results in brittle and inaccurate controllers, which have to be manually tuned for deployment.
Reinforcement learning (RL) methods have been demonstrated to be capable of learning continuous robot controllers from interactions with the environment, even for problems that include friction and contacts.
In this paper, we study how we can solve difficult control problems in the real world by decomposing them into a part that is solved efficiently by conventional feedback control methods, and the residual which is solved with RL. 
The final control policy is a superposition of both control signals.
We demonstrate our approach by training an agent to successfully perform a real-world block assembly task involving contacts and unstable objects.
\end{abstract}


\input{texs/01_introduction.tex}
\input{texs/03_background.tex}
\input{texs/04_method.tex}
\input{texs/05_experiments.tex}
\input{texs/06_results.tex}
\input{texs/07_discussion.tex}


{\small
\bibliographystyle{IEEEtran}
\bibliography{example}
}

\end{document}

%% file: texs/01_introduction.tex
\section{Introduction}\label{sec:introduction}
Robots in today's manufacturing environments typically perform repetitive tasks, and often lack the ability to handle variability and uncertainty. 
Commonly used control algorithms, such as PID regulators and the computed torque method, usually follow predefined trajectories with little adaptive behavior.
Many manufacturing tasks require some degree of adaptability or feedback to the environment, but significant engineering effort and expertise is required to design feedback control algorithms for these industrial robots.
The engineering time for fine tuning such a controller might be similar in cost to the robot hardware itself.
Being able to quickly and easily design feedback controllers for industrial robots would significantly broaden the space of manufacturing tasks that can be automated by robots.

Why is designing a feedback controller for many tasks hard with classical methods? While conventional feedback control methods can solve tasks such as path following efficiently, applications that involve contacts between the robot and its environment are difficult to approach with conventional control methods.
Identifying and characterizing contacts and friction is difficult---even if a physical model provides reasonable contact behavior, identifying the physical parameters of a contact interaction accurately is very hard.
Hence, it is often difficult to achieve adaptable yet robust control behavior, and significant control tuning effort is required as soon as these elements are introduced.
Another drawback of conventional control methods is their lack of behavior generalization.
Thus, all possible system behaviors must be considered a priori at design time.

Reinforcement learning (RL) methods hold the promise of solving these challenges because they allow agents to learn behaviors through interaction with their surrounding environments and ideally generalize to new scenarios that differ from the specifications at the control design stage.
Moreover, RL can handle control problems that are difficult to approach with conventional controllers because the control goal can be specified indirectly as a term in a reward function and not explicitly as the result of a control action.  
All of these aspects are considered enablers for truly autonomous manufacturing systems and important for fully flexible lot-size one manufacturing \cite{lotsizeone}.
However, standard RL methods require the robot learn through interaction, which can be unsafe initially, and collecting the amount of interaction that is needed to learn a complex skill from scratch can be time consuming.

\begin{figure*}[t]
    \vspace{6pt}
    \centering
    \begin{subfigure}[b]{0.45\linewidth}
        \includegraphics[width=0.9\linewidth]{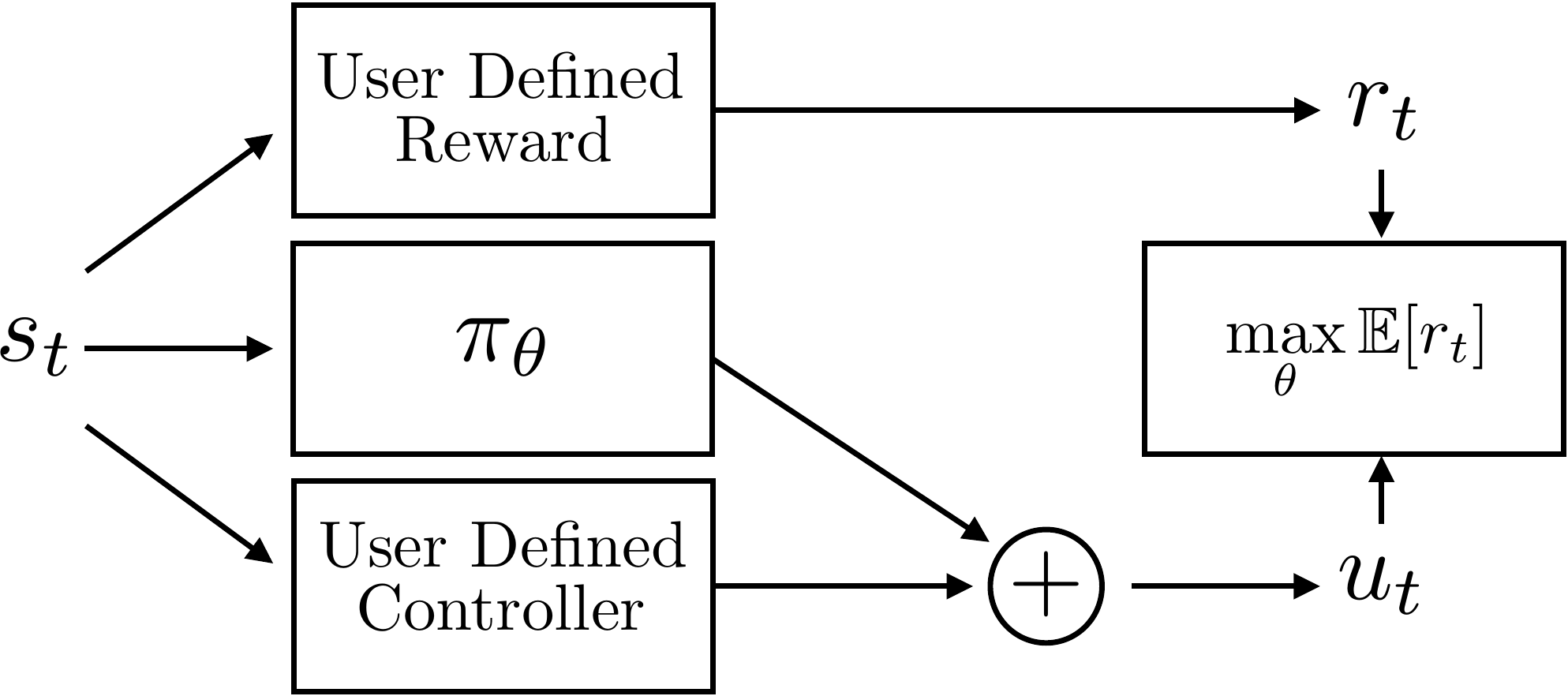}
    \end{subfigure}
    \begin{subfigure}[b]{0.54\linewidth}
        \includegraphics[width=0.99\linewidth]{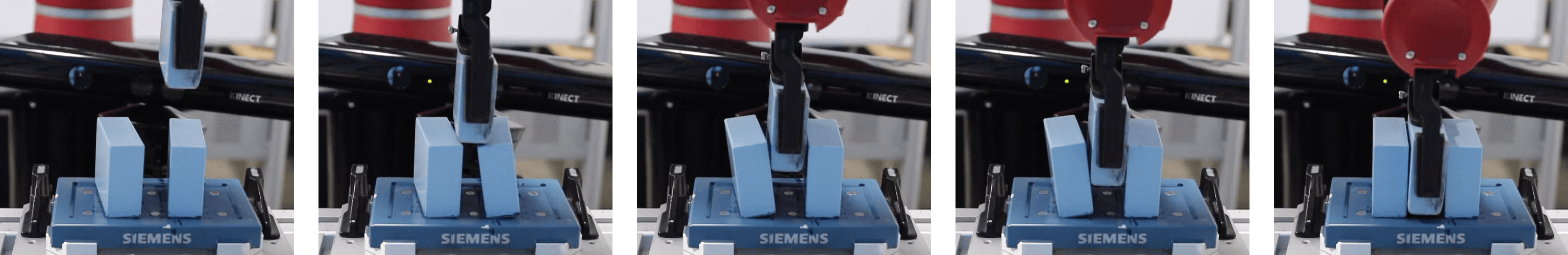} \\ \\
        \includegraphics[width=0.99\linewidth]{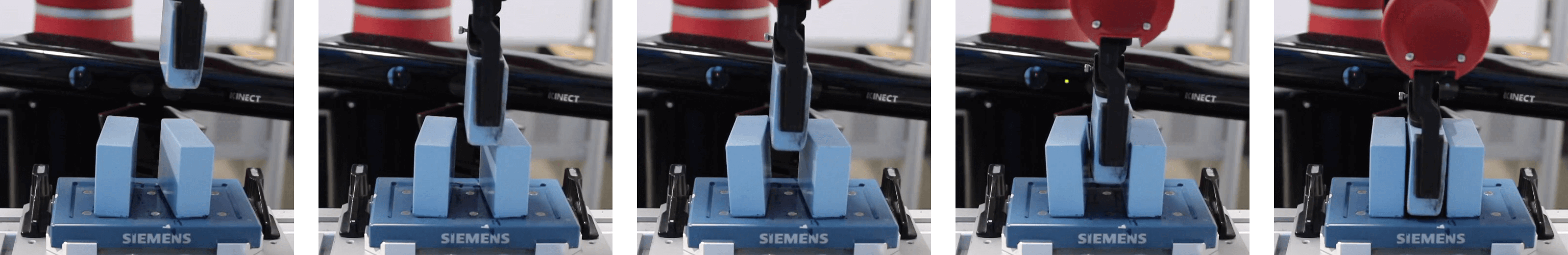}
    \end{subfigure}
    \caption{We train an agent directly in the real world to solve a model assembly task involving contacts and unstable objects. An outline of our method, which consists of combining hand-engineered controllers with a residual RL controller, is shown on the left. Rollouts of residual RL solving the block insertion task are shown on the right. Residual RL is capable of learning a feedback controller that adapts to variations in the orientations of the standing blocks and successfully completes the task of inserting a block between them. Videos are available at \href{http://residualrl.github.io}{\texttt{residualrl.github.io}}}.
    \label{fig:fig1}
\end{figure*}

In this paper, we study control problems that are difficult to approach with conventional feedback control methods. 
However, the problems possess structure that can be partially handled with conventional feedback control, e.g. with impedance control.
The residual part of the control task, which is the part that must consider contacts and external object dynamics, is solved with RL. The outputs of the conventional controller and RL are superposed to form the commanded control.
The main contribution of this paper is a methodology that combines conventional feedback control with deep RL methods and is illustrated in Fig. \ref{fig:fig1}.
Our main motivation is a control approach that is suitable for real-world control problems in manufacturing, where the exploratory behavior of RL is a safety concern and the data requirements of deep RL can be expensive.
We provide a thorough evaluation of our method on a block assembly task in simulation and on physical hardware. When the initial orientation of the blocks is noisy, our hand-designed controller fails to solve the task, while residual RL successfully learns to perform the task in under 3 hours. This suggests that we could usefully apply our method to practical manufacturing problems.

%% file: texs/03_background.tex
\section{Preliminaries}\label{sec:background}
In this section, we set up our problem and summarize the foundations of classical control and reinforcement learning that we build on in our approach.

\subsection{Problem Statement - System Theoretic Interpretation}
The class of control problems that we are dealing with in this paper can be viewed from a dynamical systems point of view as follows.
Consider a dynamical system that consists of a fully actuated robot and underactuated objects in the robot's environment.
The robot and the objects in its environment are described by their states $s_\text{m}$ and $s_\text{o}$, respectively. 
The robot can be controlled through the control input $u$ while the objects cannot be directly controlled.
However, the robot's states are coupled with the objects' states so that indirect control of $s_\text{o}$ is possible through $u$.
This is for example the case if the agent has large inertia and is interacting with small parts as is common in manufacturing.
The states of agent and objects can either be fully observable or they can be estimated from measurements.

The time-discrete equations of motion of the overall dynamical system comprise the robot and objects and can be stated as
\begin{equation}
\label{eq:eom}
    s_{t+1}=\begin{bmatrix}s_{\text{m},t+1} \\ s_{\text{o}, t+1}\end{bmatrix}
    =\begin{bmatrix}A(s_{\text{m},t}) & 0 \\ B(s_{\text{m},t}, s_{\text{o},t}) & C(s_{\text{o},t})\end{bmatrix}\begin{bmatrix}s_{\text{m},t} \\ s_{\text{o},t}\end{bmatrix}  + D\begin{bmatrix}u_t \\ 0\end{bmatrix},
\end{equation}
where the states can also be subject to algebraic constraints, which we do no state explicitly here.

The type of control objectives that we are interested in can be summarized as controlling the agent in order to manipulate the objects while also fulfilling a geometric objective such as trajectory following.
It is difficult to solve the control problem directly with conventional feedback control approaches, which compute the difference between a desired and a measured state variable.
In order to achieve best system performance feedback control methods require well understood and modeled state transition dynamics. 
Finding the optimal control parameters can be difficult or even impossible if the system dynamics are not fully known.

In \eqref{eq:eom} the state transition matrices although $A(s_\text{m})$ and $C(s_\text{o})$ are usually known to a certain extent, because they represent rigid body dynamics, the coupling matrix $B(s_\text{m}, s_\text{o})$ is usually not known. 
Physical interactions such as contacts and friction forces are the dominant effects that $B(s_\text{m}, s_\text{o})$  needs to capture, which also applies to algebraic constraints, which are functions of $s_\text{m}$ and $s_\text{o}$ as well. 
Hence, conventional feedback control synthesis for determining $u$ to control $s_\text{o}$ is very difficult, and requires trial and error in practice.
Another difficulty for directly designing feedback controllers is due to the fact that, for many control objectives, the states $s_\text{o}$ need to fulfill conditions that cannot be expressed as deviations (errors) from desired states. 
This is often the case when we only know the final goal rather than a full trajectory.

Instead of directly designing a feedback control system, we can instead specify the goal via a reward function. These reward functions can depend on both $s_\text{m}$ and $s_\text{o}$, where the terms that depend on $s_\text{m}$ are position related objectives.

\subsection{Interpretation as a Reinforcement Learning Problem}
 
In reinforcement learning, we consider the standard Markov decision process framework for picking optimal actions to maximize rewards over discrete timesteps in an environment $E$. At every timestep $t$, an agent is in a state $s_t$, takes an action $u_t$, receives a reward $r_t$, and $E$ evolves to state $s_{t+1}$. In reinforcement learning, the agent must learn a policy $u_t = \pi(s_t)$ to maximize expected returns.  We denote the return by $R_t = \sum_{i=t}^T \gamma^{(i - t)} r_i$, where $T$ is the horizon that the agent optimizes over and $\gamma$ is a discount factor for future rewards. The agent's objective is to maximize expected return from the start distribution $J = \E_{r_i, s_i \sim E, a_i \sim \pi}[R_0]$. 

Unlike the previous section, RL does not attempt to model the unknown coupled dynamics of the agent and the object. Instead, it finds actions that maximizes rewards, without making any assumptions about the system dynamics.
In this paper, we use value-based RL methods. These methods estimate the state-action value function:
\begin{align}
    Q^\pi & (s_t, u_t) = \E_{r_i, s_i \sim E, u_i \sim \pi}[R_t|s_t, u_t] \label{eqn:Q} \\
                   & = \E_{r_t, s_{t+1} \sim E}[r_t + \gamma \E_{u_{t+1} \sim \pi}[Q^\pi(s_{t+1}, u_{t+1})]] \label{eqn:bellman}
\end{align}

\noindent Equation \ref{eqn:bellman} is a recursive version of Equation \ref{eqn:Q}, and is known as the Bellman equation. The Bellman equation allows us to estimate $Q$ via approximate dynamic programming. Value-based methods can be learned off-policy, making them very sample efficient, which is vital for real-world robot learning.

%% file: texs/04_method.tex
\section{Method}\label{sec:method}

Based on the analysis in Sec. \ref{sec:background}, we introduce a control system that consists of two parts. The first part is based on conventional feedback control theory and maximize all reward terms that are functions of $s_\text{m}$. An RL method is superposed and maximizes the reward terms that are functions of $s_\text{o}$.

\subsection{Residual Reinforcement Learning}
In most robotics tasks, we consider rewards of the form:
\begin{equation}\label{eq:reward}
    r_t = f(s_\text{m}) + g(s_\text{o}).
\end{equation}
The term $f(s_\text{m})$ is assumed to be a function, which represents a geometric relationship of robot states, such as a Euclidean distance or a desired trajectory.
The second term of the sum $g(s_\text{o})$ can be a general class of functions. Concretely, in our model assembly task, $f(s_m)$ is the reward for moving the robot gripper between the standing blocks, while $g(s_o)$ is the reward for keeping the standing blocks upright and in their original positions.

The key insight of residual RL is that in many tasks, $f(s_\text{m})$ can be easily optimized a priori of any environment interaction by conventional controllers, while $g(s_\text{o})$ may be easier to learn with RL which can learn fine-grained hand-engineered feedback controllers even with friction and contacts.
To take advantage of the efficiency of conventional controllers but also the flexbility of RL, we choose:
\begin{equation}\label{eq:ctrl_seq}
    u = \pi_H(s_\text{m}) + \pi_\theta(s_\text{m}, s_\text{o})
\end{equation}
as the control action, where $\pi_H(s_\text{m})$ is the human-designed controller and $\pi_\theta(s_\text{m}, s_\text{o})$ is a learned policy parametrized by $\theta$ and optimized by an RL algorithm to maximize expected returns on the task.

Inserting \eqref{eq:ctrl_seq} into \eqref{eq:eom} one can see that a properly designed feedback control law for $\pi_H(s_\text{m})$ is able to provide exponentially stable error dynamics of $s_\text{m}$ if the learned controller $\pi_\theta$ is neglected and the sub statespace is stabilizable.
This is equivalent to maximizing \eqref{eq:reward} for the case $f$ represents errors between actual and desired states.

The residual controller $\pi_\theta(s_\text{m}, s_\text{o})$ can now be used to maximize the reward term $g(s_\text{o})$ in \eqref{eq:reward}.
Since the control sequence \eqref{eq:ctrl_seq} enters \eqref{eq:eom} through the dynamics of $s_\text{m}$ and $s_\text{m}$ is in fact the control input to the dynamics of $s_\text{o}$, we cannot simply use the a-priori hand-engineered feedback controller to achieve zero error of $s_\text{m}$ and independently achieve the control objective on $s_\text{o}$.
Through the coupling of states we need to perform an overall optimization of \eqref{eq:ctrl_seq}, whereby the hand-engineered feedback controller provides internal structures and eases the optimization related to the reward term $f(s_\text{m})$.

\setlength{\textfloatsep}{0.09cm}
\begin{algorithm}
   	\caption{Residual reinforcement learning}
   	\label{alg:residualrl}
   	\begin{algorithmic}[1]
    \REQUIRE policy $\pi_\theta$, hand-engineered controller $\pi_\text{H}$.
    \FOR{$n=0,...,N-1$ episodes}
        \STATE Initialize random process $\mathcal{N}$ for exploration
        \STATE Sample initial state $s_0 \sim E$.
        \FOR{$t=0,...,H -1$ steps}
            \STATE Get policy action $u_t = \pi_\theta(s_t) + \mathcal{N}_t$.
            \STATE Get action to execute $u'_t = u_t + \pi_\text{H}(s_t)$.
            \STATE Get next state $s_{t+1} \sim p(\cdot \mid s_t, u'_t)$.
            \STATE Store $(s_t, u_t, s_{t+1})$ into replay buffer $\mathcal R$.
            \STATE Sample set of transitions $(s, u, s') \sim \mathcal R$.
            \STATE Optimize $\theta$ using RL with sampled transitions.
        \ENDFOR
    \ENDFOR
   	\end{algorithmic}
\end{algorithm}
\setlength{\floatsep}{0.09cm}
\subsection{Method Summary}
Our method is summarized in Algorithm \ref{alg:residualrl}. The key idea is to combine the flexibility of RL with the efficiency of conventional controllers by additively combining a learnable parametrized policy with a fixed hand-engineered controller.

\begin{figure*}[t]
    \vspace{6pt}
    \centering
    \includegraphics[height=1.1in]{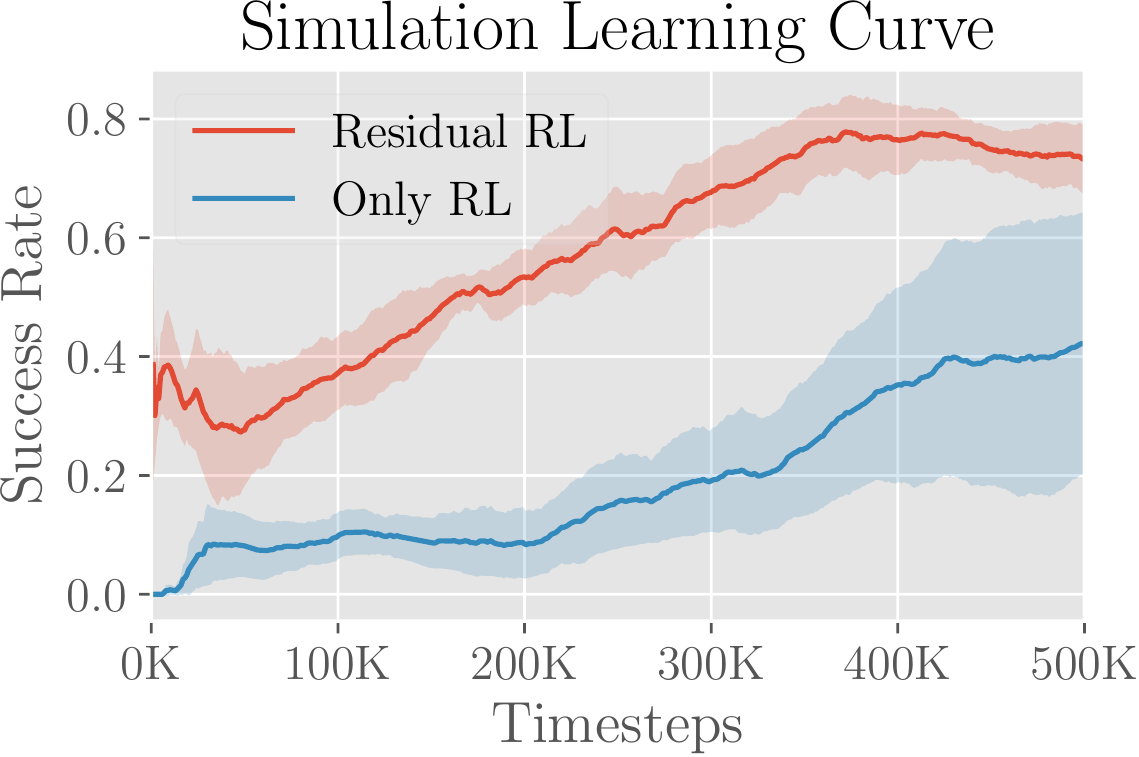}
    \includegraphics[height=1.1in]{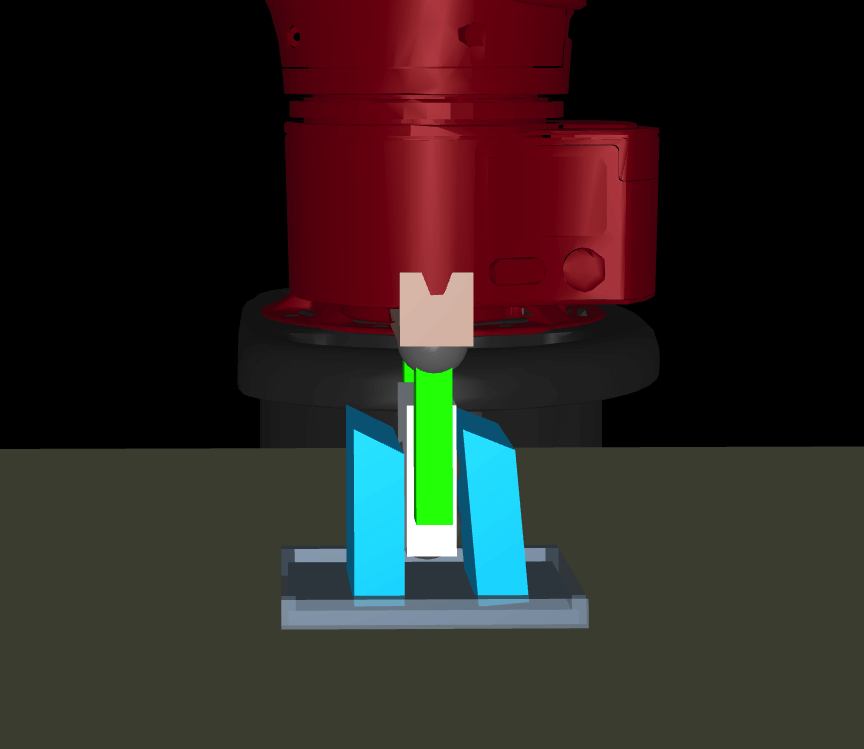} 
    \hspace{1cm}
    \includegraphics[height=1.1in]{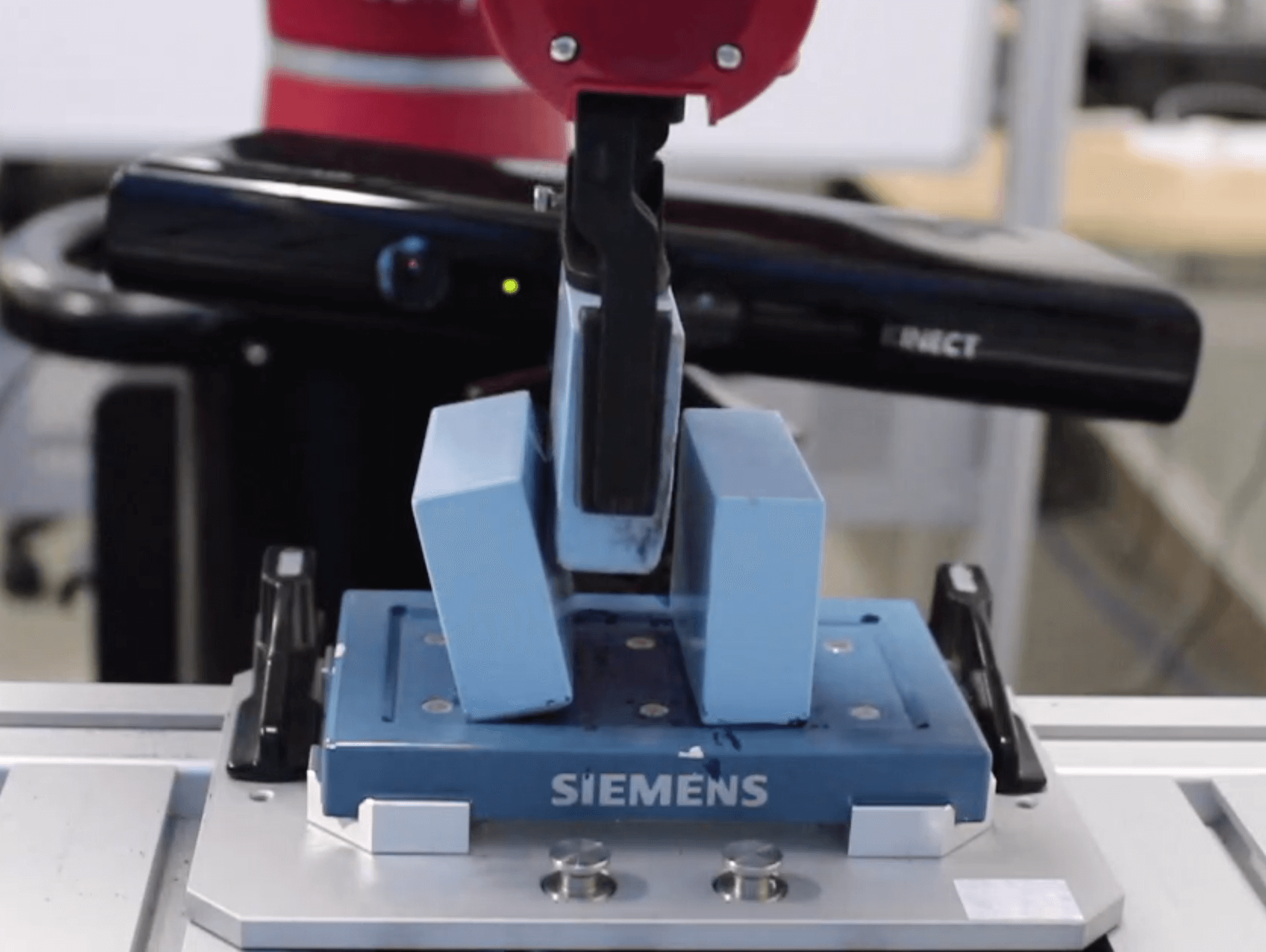} 
    \vspace{0.1cm}
    \includegraphics[height=1.1in]{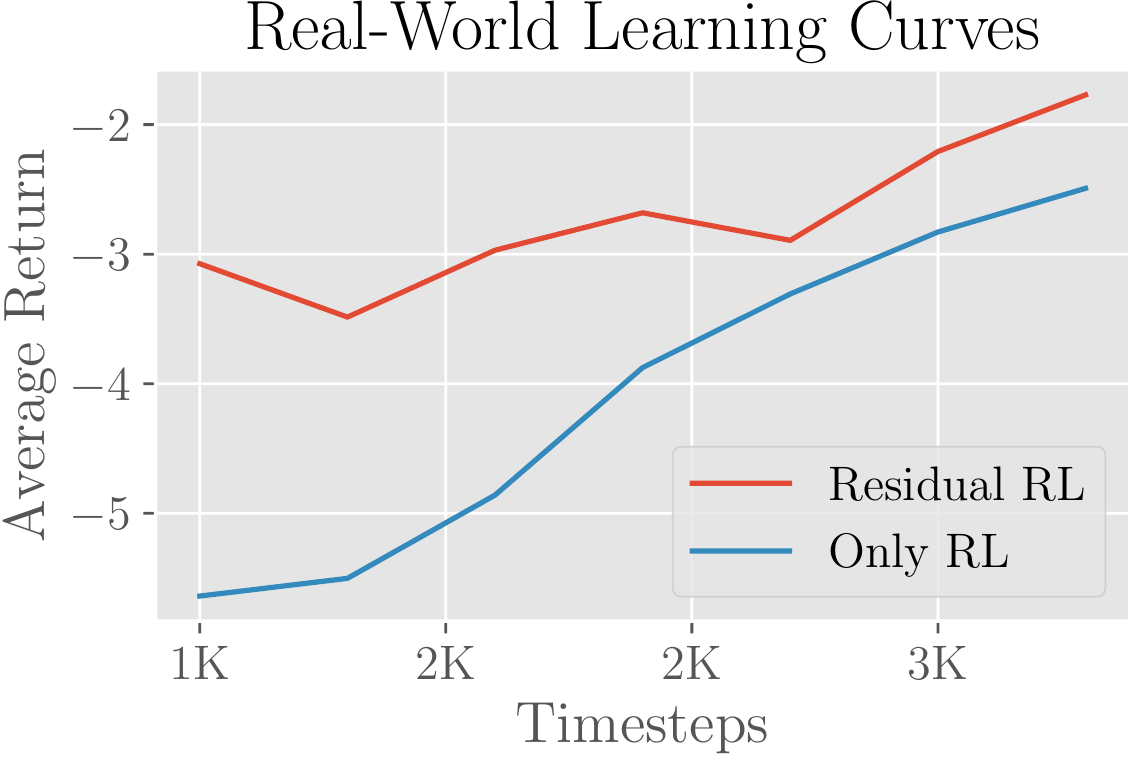}
    \caption{Block assembly task in simulation (left) and  real-world (right). The task is to insert a block between the two blocks on the table without moving the blocks or tipping them over. In the learning curves, we compare our method with RL without any hand-engineered controller\protect\footnotemark. In both simulation and real-world experiments, we see that residual RL learns faster than RL alone, while achieving better performance than the hand-engineered controller. }%
    \label{fig:fig2}
\end{figure*}

As our underlying RL algorithm, we use a variant of twin delayed deep deterministic policy gradients (TD3)
as described in~\cite{fujimoto2018td3}. TD3 is a value-based RL algorithm for continuous control based off of the deep deterministic policy gradient (DDPG) algorithm \cite{lillicrap2015continuous}. We have found that TD3 is stable, sample-efficient, and requires little manual tuning compared to DDPG.
We used the publicly available \href{https://github.com/vitchyr/rlkit}{\texttt{rlkit}} implementation of TD3 \cite{pong2018tdm}. Our method is independent of the choice of RL algorithm, and we could apply residual RL to any other RL algorithm.

%% file: texs/05_experiments.tex
\section{Experimental Setup}\label{sec:experiments}

We evaluate our method on the task shown in Fig. \ref{fig:fig2}, both in simulation and in the real world. This section introduces the details of the experimental setup and provides an overview of the experiments.

\subsection{Simulated Environment}
We use MuJoCo~\cite{todorov12mujoco}, a full-featured simulator for model-based optimization considering body contacts, to evaluate our method in simulation.
This environment consists of a simulated Sawyer robot arm with seven degrees of freedom and a parallel gripper. 
We command the robot with a Cartesian-space position controller.
Two blocks each with 3-DOF and one angled corner on the top are placed on a defined platform on the table in front of the robot. 
To allow block insertion between the standing blocks, a sufficiently large gap is defined (this gap represents the goal position). 
Both standing blocks can slide in $x$- and $y$-direction and topple around the $y$-axis.
The agent receives the end effector position, the end effector forces and torques in relation to the $x$-, $y$- and $z$-axes, all block positions, and the goal position as the observation. In the robot's initial position one block for the insertion process is already picked up by the gripper claws and the gripper is located above the blocks on the tabletop. 

We use a reward function 
    \begin{align}
        r_t = - \|x_g - x_t \|_2 - \lambda (\|\theta_{l} \|_1 + \|\theta_{r} \|_1)
    \end{align}
where $x_t$ is the current block position, $x_g$ is the goal position, $\theta_{l}$, $\theta_{r}$ are the angles with respect to the table (in the y-axis) of the left and right blocks, respectively. $\lambda$ is a hyperparameter.

\subsection{Real-World Environment}
The real-world environment is largely the same as the simulated environment, except for the controller, rewards, and observations.
We command the robot with a compliant joint-space impedance controller we have developed to be smooth and tolerant of contacts.
The positioning of the block being inserted is similar to the simulation but the observation the agent receives is different. Instead of receiving ground truth position information, it is estimated from a camera-based tracking system.
Due to the blocks' slight weight and their capability of sliding in the plane ($x$, $y$), the Sawyer is not able to measure contact forces regarding these axes.
Therefore, we only add the end effector forces in $z$-direction to the observation space instead of observing the end effector forces and torques regarding to the $x$-, $y$- and $z$-axes.
The reward function was slightly different, being defined as: 
    \begin{align}
    \begin{split}
    r_t = - \|x_g - x_t \|_2 - \lambda (\|\theta_{l} \|_1 + \|\theta_{r} \|_1) \\ - \mu \|X_g - X_t \|_2 - \beta (\|\phi_{l} \|_1 + \|\phi_{r} \|_1)
    \end{split}
    \end{align}
where $x_t$ is the current end effector position, $x_g$ is the goal position, $X_t$ describes the current position of both standing blocks, $X_g$ their desired positions, $\theta_{l}$, $\theta_{r}$ are the angles of the current orientation with respect to the table (in the y-axis) and $\phi_{l}$ and $\phi_{r}$ are the angles of the current orientation with respect to the z-axis of the left and the right block respectively $\lambda$, $\mu$, and $\beta$ are hyperparameters. 

%

\footnotetext{In all simulation plots, we use 10 random seeds and report a $95\%$ confidence interval for the mean.}

\subsection{Overview of Experiments}
%
In our experiments we evaluate the following research questions:
\begin{enumerate}
    \item Does incorporating a hand-designed controller improve the performance and sample-efficiency of RL algorithms, while still being able to recover from an imperfect hand-designed controller?
    \item Can our method allow robots to be more tolerant of variation in the environment?
    \item Can our method successfully control noisy systems, compared to classical control methods?
\end{enumerate}

\section{Experiments}

\subsection{Sample Efficiency of Residual RL}
In this section, we compare our residual RL method with the human controller alone and RL alone. The following methods are compared:
\begin{enumerate}
    \item Only RL: using the same underlying RL algorithm as our method but without adding a hand-engineered policy
    \item Residual RL: our method which trains a superposition of the hand-engineered controller and a neural network policy, with RL
\end{enumerate}

\subsection{Effect of Environment Variation}
In automation, environments can be subject to noise and solving manufacturing tasks become more difficult as variability in the environment increases. 
It is difficult to manually design feedback controllers that are robust to environment variation, as it might require significant human expertise and tuning. In this experiment, we vary the initial orientation of the blocks during each episode and demonstrate that residual RL can still solve the task. We compare its performance to that of the hand-engineered controller.

To introduce variability in simulation, on every reset we sampled the rotation of each block independently from a uniform distribution $U[-r, r], r \in \{0, 0.05, 0.1, 0.15, 0.2, 0.25, 0.3\}$.

Similarly, in the real world experiments, on every reset we randomly rotated each block to one of three orientations: straight, tilted clockwise, or tilted counterclockwise (tilt was $\pm$  $20^{\circ}$ from original position).

\subsection{Recovering from Control Noise}

Due to a host of issues, such as defective hardware or poorly tuned controller parameters, feedback controllers might have induced noise. Conventional feedback control policies are determined a priori and do not adapt their behavior from data.
However, RL methods are known for their ability to cope with shifting noise distributions and are capable of recovering from such issues.

\begin{figure*}[t]
    \vspace{6pt}
    \centering
    \begin{subfigure}[b]{0.38\linewidth}
        \begin{tabular}{ | c || c | c |}
            \hline
            Misaligned? & No & Yes  \\ \hline
            Residual RL & 20/20 & 15/20  \\ \hline
            Hand-engineered  & 20/20 & 2/20  \\ \hline
        \end{tabular}
        \vspace{1.2cm} \\
        \centering
        (a)
    \end{subfigure}
    \begin{subfigure}[b]{0.3\linewidth}
        \includegraphics[width=0.99\linewidth]{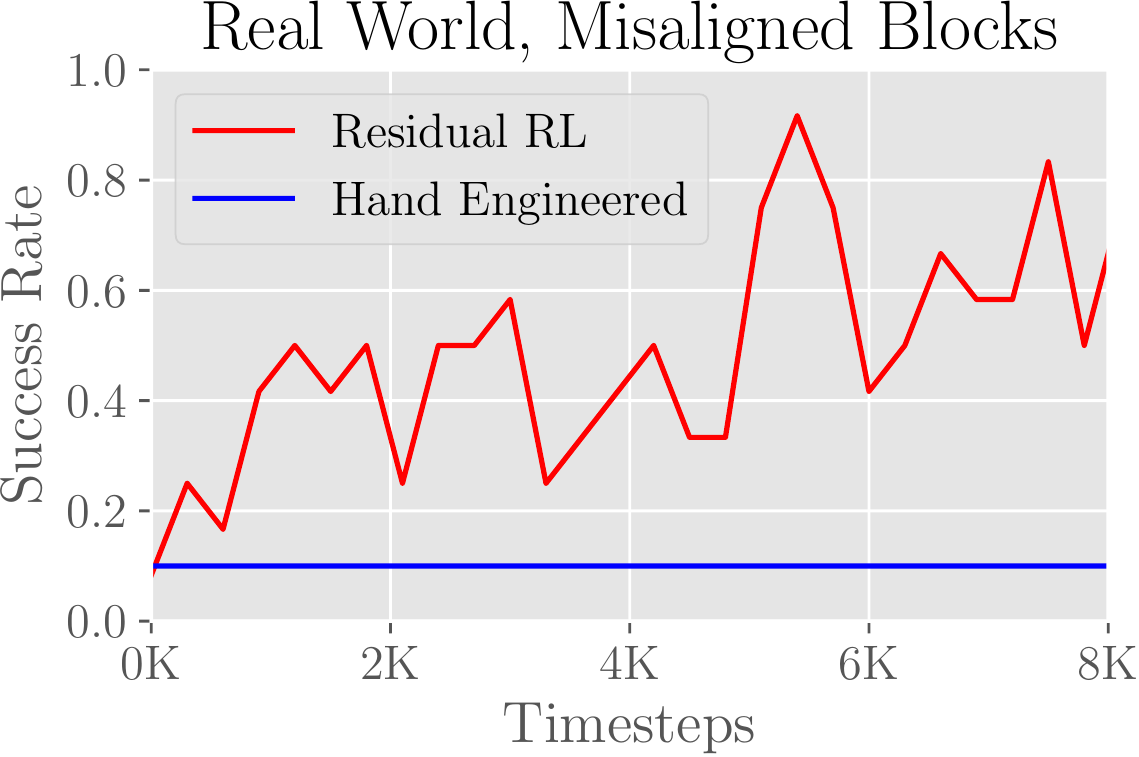} \\
        \centering
        (b)
    \end{subfigure}
    \begin{subfigure}[b]{0.3\linewidth}
        \includegraphics[width=0.99\linewidth]{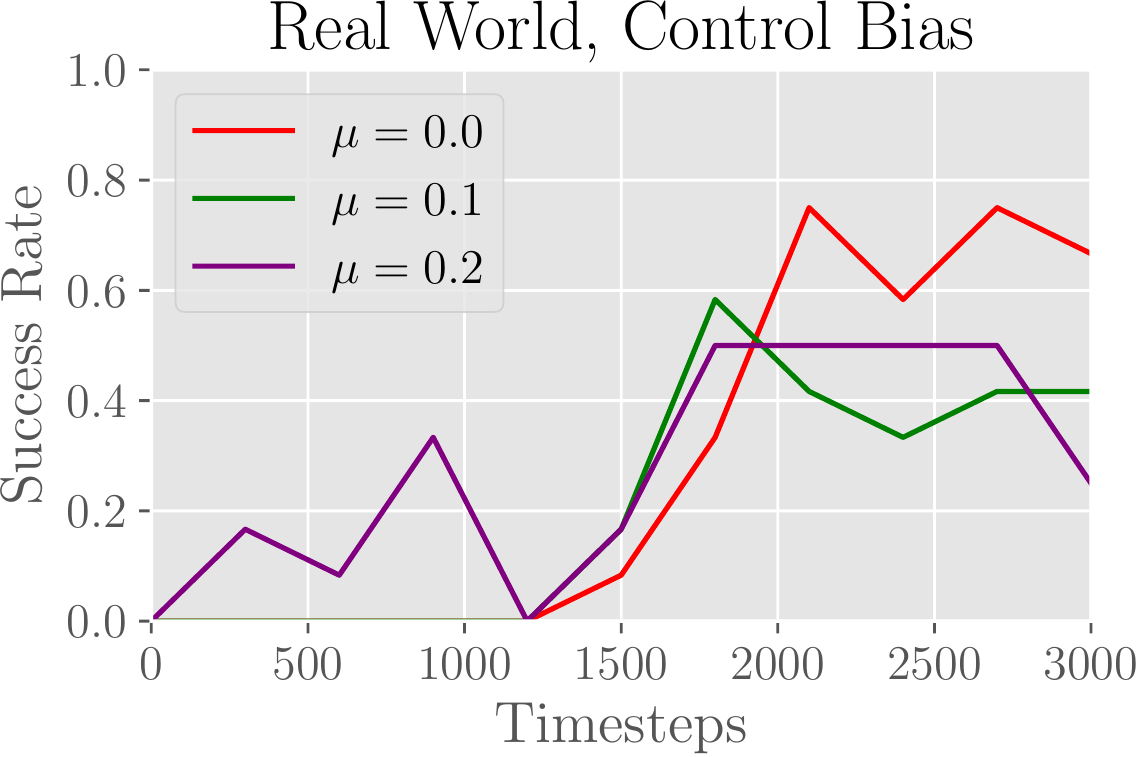} \\
        \centering
        (c)
    \end{subfigure}
    \vspace{0.2cm}
    \caption{Outcome of our residual RL method in different experiments during the block assembly task in the real-world. Success rate is recorded manually by human judgment of whether the blocks stayed upright and ended in the correct position. Plot (a) compares the insertion success of residual RL and hand-designed controller depending on the block orientation during run time. Plot (b) shows the success rate of the insertion process during training, where on every reset the blocks are randomly rotated: straight, tilted clockwise, or tilted counterclockwise ($\pm$  $20^{\circ}$) and plot (c) shows the increasing success rate of our method for biased controllers as well even as control bias increases.}%
    \label{fig:environment_variation}
\end{figure*}

In this experiment, we introduce a control noise, including biased control noise, and demonstrate that residual RL can still successfully solve the task, while a hand-engineered controller cannot. The control noise follows a normal distribution and is added to the control output of the system at every step:
\begin{align}
    u'_t = u_t + \mathcal{N}(\mu, \sigma^2)
\end{align}

To test tolerance to control noise, we set $\mu = 0$ and vary $\sigma \in [0.01, 0.1]$. In theory, RL could adapt to a noisy controller by learning more robust solutions to the task which are less sensitive to perturbations.

Furthermore, to test tolerance to a biased controller, we set $\sigma = 0.05$ and vary $\mu \in [0, 0.2]$. To optimize the task reward, RL can learn to simply counteract the bias. 

\subsection{Sim-to-Real with Residual RL}

As an alternative to analytic solutions of real-world control problems, we can often instead model the forward dynamics of the problem (ie. a simulator). With access to such a model, we can first find a solution to the problem with our possibly inaccurate model, and then use residual RL to find a realistic control solution in the real world.

In this experiment, we attempt the block insertion task with the side blocks fixed in place. The hand-engineered policy $\pi_\text{H}$ in this case comes from training a parametric policy in simulation of the same scenario (with deep RL). We then use this policy as initialization for residual RL in the real world.

%% file: texs/06_results.tex
\section{Results}\label{sec:results}

We trained our method to optimize the insertion process in simulation as well as on physical hardware.
This section provides the results of our discussed experiments and shows the functionality of our method.

\subsection{Sample Efficiency of Residual RL}

First, we compare residual RL and pure RL without a hand-engineered controller on the insertion task. 
Fig. \ref{fig:fig2} shows in simulation and real-world that residual RL achieves a better final performance and requires less samples than RL alone, both in simulation and on physical hardware. 
Unlike residual RL, the pure RL approach needs to learn the structure of the position control problem from scratch, which explains the difference in sample efficiency.
As samples are expensive to collect in the real world, residual RL is better suited for solving real-world tasks. Moreover, RL shows a broader spatial variance during training and needs to explore a wider set of states compared to residual RL, which can be potentially dangerous in hardware deployments.

\begin{figure*}[t]
    \vspace{6pt}
    \centering
    \begin{subfigure}[b]{0.32\linewidth}
        \includegraphics[width=0.99\linewidth]{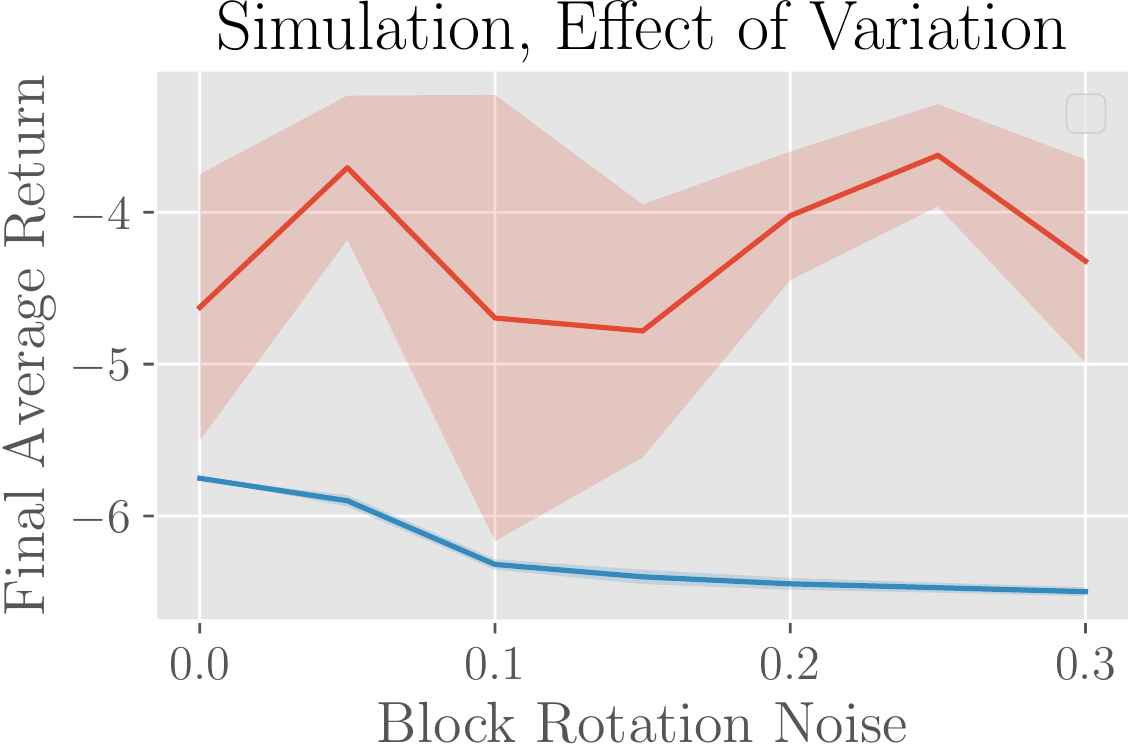} \\
        \centering
        (a)
    \end{subfigure}
    \begin{subfigure}[b]{0.32\linewidth}
        \includegraphics[width=0.99\linewidth]{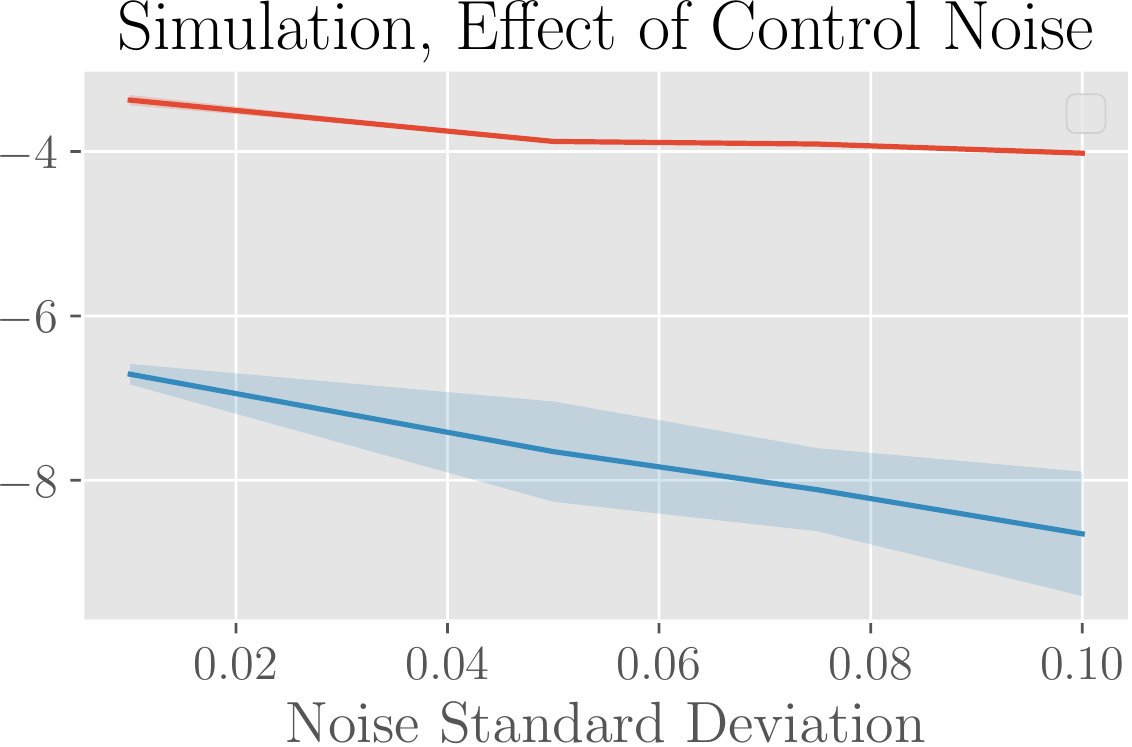} \\
        \centering
        (b)
    \end{subfigure}
    \begin{subfigure}[b]{0.32\linewidth}
        \includegraphics[width=0.99\linewidth]{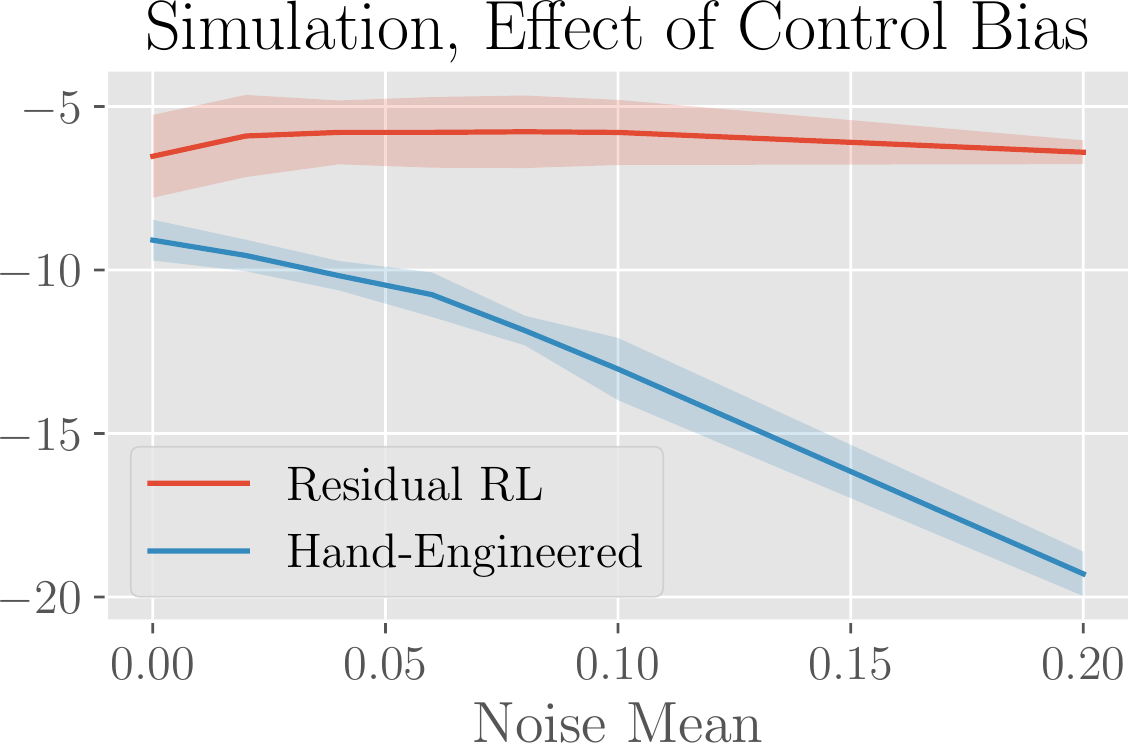} \\
        \centering
        (c)
    \end{subfigure}
    \caption{Simulation results for different experiments. In each plot, the final average return obtained by running the method for various settings of a parameter is shown. Plot (a) shows that residual RL can adjust to noise in the environment caused by rotation of the blocks in a range of $0$ to $0.3$\,rad. In plot (b), residual RL finds robust strategies in order to reduce the effect of control noise, as the final average return is not greatly affected by the magnitude of noise. Plot (c) shows that residual RL can compensate for biased controllers and maintains good performance as control bias increases, while the performance of the hand-designed controller dramatically deteriorates with higher control bias.} %
    \label{fig:control_bias}
\end{figure*}

\subsection{Effect of Environment Variation}

In previous set of experiments, both standing blocks were placed in their initial position without any position or orientation error. 
In this case, the hand-engineered controller performs well, as both blocks are placed such that there is a sufficiently large defined gap for insertion. 
However, once the initial orientation of the blocks is randomized, the gap between the blocks and the goal position does not afford directly inserting from above.
Therefore, the hand-engineered controller struggles to solve the insertion task, succeeding in only 2/20 trials, while residual RL still succeeds in 15/20 trials. These results are summarized in Fig. \ref{fig:environment_variation}\,(a). Rollouts from the learned policy are included in \ref{fig:fig1}. In this experiment, the agent demonstrably learns consistent small corrective feedback behaviors in order to slightly nudge the blocks in the right direction without tipping them over, a behavior that is very difficult to manually specify.

The result of this experiment showcases the strength of residual RL. Since the human controller specifies the general trajectory of the optimal policy, environment samples are required only to learn this corrective feedback behavior. The real-world learning curve for the experiment in Fig. \ref{fig:environment_variation}\,(b) shows that this behavior is gradually acquired over the course of eight thousand samples, which is only about three hours of real-world training time.

We further studied the effect of the block orientation changing after every reset in simulation. The results are shown in Fig. \ref{fig:control_bias}\,(a). 
The simulation results show that the performance of the hand-engineered controller decreases as the block rotation angle increases, whereas our control method maintains a constant average performance over different variations. 

\begin{figure}[b]
    \vspace{6pt}
    \centering
    \includegraphics[width=0.99\linewidth]{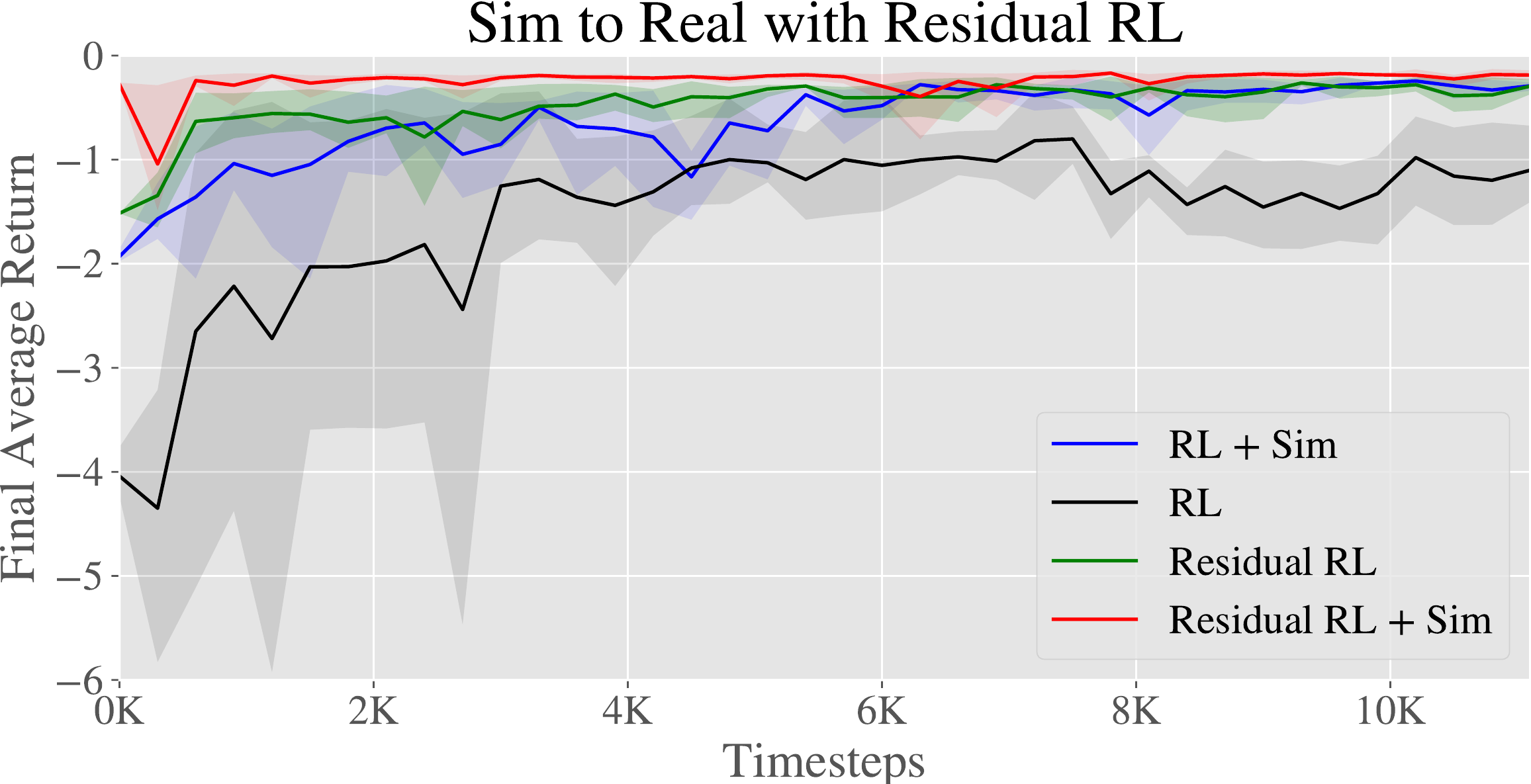} \\
    \centering
    \caption{Real-world block insertion results using residual RL for sim-to-real transfer. ``Sim'' indicates that the real-world policy was initialized by reinforcement learning in simulation. Residual RL with simulation initialization successfully solves the task with little environment experience required. } %
    \label{fig:sim_to_real}
\end{figure}

\subsection{Recovering from Control Noise}

In this experiment, we observe that residual RL is able to cope with actuator noise, including biased actuator noise.
Quantitative results for simulation are shown in  Fig.~\ref{fig:control_bias}\,(b) and (c). In Fig.~\ref{fig:control_bias}\,(c) our method keeps the average return constant and correct for biased controllers even as control bias increases, whereas the hand-engineered controller cannot compensate biased input and its performance deteriorates as control bias increases.
The same applies for adding control noise to the control output as shown in Fig.~\ref{fig:control_bias}\,(b).

For the hardware experiments, only biased actuator noise is investigated. These results are shown in Fig.~\ref{fig:environment_variation}\,(c). These learning curves show that even as more control bias is introduced, training in the real world proceeds without significant issues. This result suggests the potential for RL to address practical issues in automation such as sensor drift.

\subsection{Sim-to-Real with Residual RL}

The result of the sim-to-real experiment is shown in Fig. \ref{fig:sim_to_real}. In this experiment, each setting was run with three random seeds. Adding policy initialization from simulation significantly speeds up both RL and residual RL. In particular, residual RL with policy initialization from simulation successfully solves the task extremely quickly: in under one thousand timesteps of environment interaction. This method poses a highly sample efficient, practical way to solve robotics problems with  difficult contact dynamics.

\section{Related Work}\label{sec:related_work}

Reinforcement learning for robotics holds the promise of greater autonomy and reliability, which could be vital to improving our manufacturing processes beyond its current limitations.
RL methods have been difficult to apply in robotics because of sample efficiency, safety, and stability issues. Still, RL has been used to allow robots to learn tasks such as playing table tennis \cite{peters2010reps}, swinging up a cartpole and balancing a unicycle \cite{deisenroth2011pilco}, grasping~\cite{pinto2017robust, levine2017grasping}, opening a door \cite{Gu2016b}, and general manipulation tasks \cite{levine2016gps, haarnoja2018sac}. RL, particularly deep RL, tends to be data-hungry; even learning simple tasks can require many hours of interaction. To bring these methods into factories and warehouses, they must be able to consistently solve complex tasks, multi-step tasks. One way to enable these methods to solve these complex tasks is to introduce prior human knowledge into the learning system, as our method does.

Prior work in RL has incorporated human prior knowledge for solving tasks in various ways. One such way is reward shaping \cite{ng1999rewardshaping}, where additional rewards auxiliary to the real objective are included in order to guide the agent towards the desired behavior. Reward shaping can effectively encode a policy. For example, to train an agent to perform block stacking, each step can be encoded into the reward \cite{popov17stacking}. Often, intensive reward shaping is key for RL to succeed at a task and prior work has even considered reward shaping as part of the learning system \cite{daniel2014activereward}. Reward shaping in order to tune agent behavior is a very manual process and recovering a good policy with reward shaping can be as difficult as specifying the policy itself. Hence, in our method we allow for human specification of both rewards and policies---whichever might be more practical for a particular task.

Further work has incorporated more specialized human knowledge into RL systems. One approach is to use trajectory planners in RL in order to solve robotics tasks \cite{thomas2018cad}. However, since the method optimizes trajectory following instead of the task reward, generalization can be difficult when aspects of the environment change. Other work has focused on human feedback \cite{loftin2014discretehumanfeedback, saunders2018trialwithouterror, torrey2013teachingbudget, frank2008rareevents, christiano2017humanpreferences} to inform the agent about rewards or to encourage safety. However, in many robotics tasks, providing enough information about the task through incremental human feedback is difficult.

Another way to include prior knowledge in RL is through demonstrations~\cite{peters2008baseball, kober2008mp, rajeswaran2018dextrous, hester17dqfd, vecerik17ddpgfd, nair2018demonstrations}. Demonstrations can substantially simplify the exploration problem as the agent begins training having already received examples of high-reward transitions and therefore knows where to explore \cite{subramanian2016efd}. However, providing demonstrations requires humans to be able to teleoperate the robot to perform the task. In contrast, our method only requires a conventional controller for motion, which ships with most robots.

Prior knowledge can also be induced through neural network architecture choices. Deep residual networks with additive residual blocks achieved state of the art results in many computer vision tasks by enabling training of extremely deep networks \cite{he2016resnet}. In RL, structured control nets showed improvement on several tasks by splitting the policy into a linear module and a non-linear module \cite{srouji18structuredcontrolnets}. Most closely related to our work, residual policy learning concurrently and independently explores training a residual policy in the context of simulated long-horizon, sparse-reward tasks with environment variation and sensor noise \cite{silver18residualpolicylearning}. Our work instead focuses on achieving practical real-world training of contact-intensive tasks.

%% file: texs/07_discussion.tex
\section{Discussion and Future Work}\label{sec:discussion}

In this paper we study the combination of conventional feedback control methods with deep RL. 
We presented a control method that utilizes conventional feedback control along with RL to solve complicated manipulation tasks involving friction and contacts with unstable objects. 
We believe this approach can accelerate learning of many tasks, especially those where the control problem can be solved in large part by prior knowledge but requires some model-free reasoning to solve perfectly.
Our method extends conventional feedback control and allows us to automatically solve problems from data that would otherwise require human experts to design solutions for. 
In this sense, our method is related to differentiable programming \cite{wang2018differentiablepogramming} or automated algorithm design \cite{Sunderhauf2018}. 
Our results demonstrate that the combination of conventional feedback control and RL can circumvent the disadvantages of both and provide a sample efficient robot controller that can cope with contact dynamics.

A limitation of our method is that it requires a carefully crafted vision setup to infer positions and angles in the real world. This requirement limits its application to novel scenes, and also limits the quality of the learned feedback controller if the hand-engineered vision system does not preserve important information for the feedback controller such as the precise location of edges. Similar to compensating for control bias, the learned controller could also compensate for perception noise and bias. Therefore, one interesting direction for future work is incorporating vision input end-to-end into the learning process.

\section{Acknowledgements}

We would like to thank Murtaza Dalal for contributions to the robot controller code we built upon, as well as Vitchyr Pong and Pulkit Agrawal for useful discussions about this work. 
This work was supported by the Siemens Corporation, the Office of Naval Research under a Young Investigator Program Award, and Berkeley DeepDrive.